\documentclass[10pt,twocolumn,letterpaper]{article}

\usepackage{cvpr}
\usepackage{times}
\usepackage{epsfig}
\usepackage{graphicx}
\usepackage{amsmath}
\usepackage{amssymb}
\usepackage{array}
\newcolumntype{?}{!{\vrule width 1.3pt}}
\usepackage{multirow}
\usepackage{makecell}
 \usepackage[hyphens]{url}

\usepackage{algorithm}
\usepackage{algorithmic}

\def\eg{\emph{e.g.~}}
\def\etal{{\em et al.~}}
\def\ie{\emph{i.e.~}}

\usepackage[pagebackref=true,breaklinks=true,letterpaper=true,colorlinks,bookmarks=false]{hyperref}

  \cvprfinalcopy 

\def\cvprPaperID{3096} 

\begin{document}


\title{Empirical Upper Bound in Object Detection and More}

\author{Ali Borji \ \ \ \ \ \ \ \ \ \ \ \ \ \ \ \ \ \  Seyed Mehdi Iranmanesh$^{\dagger}$\thanks{Work done during internship at MarkableAI.} \\ 
$^\dagger$West Virginia University   \\
{\tt\small aliborji@gmail.com \ \ \ \ 
 seiranmanesh@mix.wvu.edu
}
}

\maketitle

\begin{abstract}
\vspace{-10pt}
Object detection remains as one of the most notorious open problems in computer vision. Despite large strides in accuracy in recent years, modern object detectors have started to saturate on popular benchmarks raising the question of how far we can reach with deep learning tools and tricks. Here, by employing 2 state-of-the-art object detection benchmarks, and analyzing more than 15 models over 4 large scale datasets, we I) carefully determine the upper bound in AP, which is {\bf 91.6\% on VOC} (test2007), {\bf 78.2\% on COCO} (val2017), and 58.9\% on OpenImages V4 (validation), regardless of the IOU. These numbers are much better than the mAP of the best model\footnote{The best published mAP on {\bf COCO \underline{test-dev} is 51.0} by EfficientDet~\cite{tan2019efficientdet}. See~\cite{coco_latest} for the latest results on COCO dataset.} ({\bf 47.9\% on VOC}, and {\bf 46.9\% on COCO}; IOUs=.5:.95), II) characterize the sources of errors in object detectors, in a novel and intuitive way, and find that classification error (confusion with other classes and misses) explains the largest fraction of errors and weighs more than localization and duplicate errors, and III) analyze the invariance properties of models when surrounding context of an object is removed, when an object is placed in an incongruent background, and when images are blurred or flipped vertically. We find that models generate boxes on empty regions and that context is more important for detecting small objects than larger ones. Our work taps into the tight relationship between recognition and detection and offers insights to build better models\footnote{Our code is publicly available at~\cite{ourCode}.}.

\end{abstract}

\vspace{-17pt}
\section{Introduction and Motivation}
\vspace{-7pt}

Several years of extensive research on object detection has resulted in accumulation of an overwhelming amount of knowledge regarding model backbones, tricks for model training and optimization, data collection and annotation, and model evaluation and comparison~\cite{zou2019object}, to a point that separating wheat from chaff is very difficult. As an example, truly understanding and implementing Average Precision (AP) is frustratingly difficult. A quick Google search returns numerous blogs and codes with discrepant explanations of AP. To make matters even worse, it is not quite clear whether AP has started to saturate, whether progress is significant, and more importantly how far we can improve following the current path, making one wonder maybe we have reached the peak of performance using deep learning. Further, we do not know what is holding us back from making progress in object detection, compared to human-level (although debatable) accuracy on object recognition. 

To shed light on the above matters, first we systematically and carefully approximate the empirical upper bound in AP. We hypothesize that the upper bound AP (UAP) is the score of the best recognition model that is trained on the training target bounding boxes and is then used to label the testing target boxes. We also investigate whether visual context surrounding a target object or its overlapping boxes can improve the upper bound AP. Second, we identify bottlenecks by characterising the type of errors that object detectors make and measure the impact of each one on performance. Third, we study the invariance properties of various object detectors on different types of transformations including incongruent context, scale, blur, vertical flip, etc.

\begin{figure}[t]
\vspace{-11pt}
   \includegraphics[width=\linewidth]{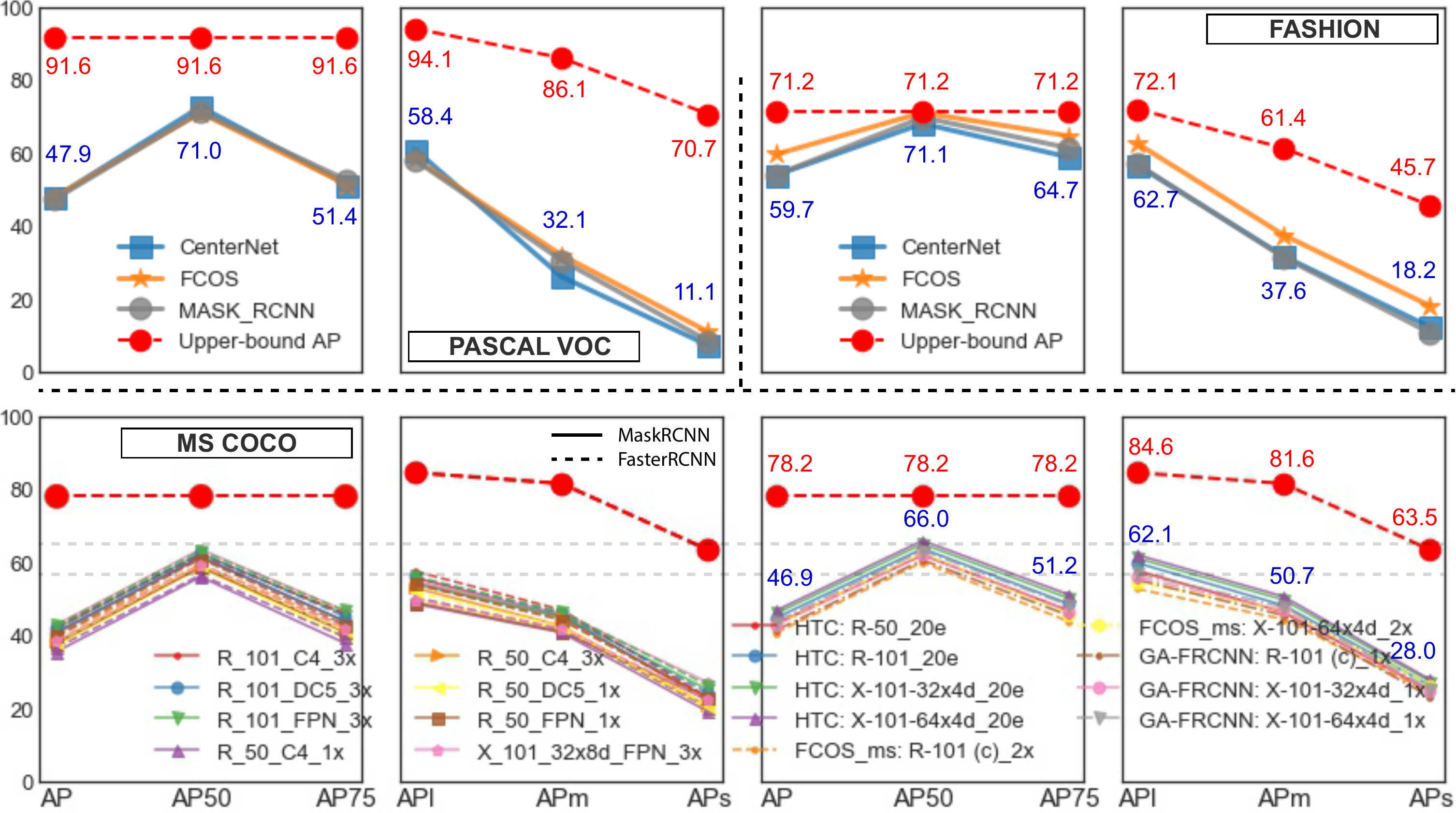}
\vspace{-13pt}
   \caption{\footnotesize{{\bf Upper bound AP} (in red) and scores of the best model (in blue; FCOS~\cite{tian2019fcos} on VOC and FASHION, and Hybrid Task Cascade~\cite{chen2019hybrid} on COCO. It shows that scale remains the major problem in object detection.}}
\vspace{-18pt}   
\label{fig:coco_fashion_voc}
\end{figure}

In a nutshell, we find that there is a large gap between the performance of the best detection models and the empirical upper bound as shown in Fig.~\ref{fig:coco_fashion_voc}. This entails that there is a hope to reach this peak with the current tools, if we can find smarter ways to adopt object recognition models for object detection. We also find that classification remains as the major bottleneck in object detection and is more critical over small objects. Specifically, object detection models inherit the main limitations of CNNs which is the lack of invariance. Example failure cases include generating a lot of bounding boxes on a white background containing a single object, and failing to detect objects in incongruent contexts, vertically flipped images or blurred ones. It seems that humans can still manage to solve these tasks, although with higher effort and lower performance than intact images.

\vspace{-5pt}
\section{Related Work}
\vspace{-5pt}
We discuss three lines of related works. The first one includes {\bf works that strive to understand detection approaches}, identify their shortcomings, and pinpoint where more research is needed.
Parikh~\etal~\cite{parikh2011finding} aimed to find the weakest links in person detectors by replacing different components in a pipeline (\eg part detection, non-maxima-suppression) with human annotations. Mottaghi~\etal~\cite{mottaghi2015human} proposed human-machine CRFs for identifying bottlenecks in scene understanding. Hoeim~\etal~\cite{hoiem2012diagnosing} inspected detection models in terms of their localization errors, confusion with other classes, and confusion with the background on PASCAL dataset. They also conducted a meta-analysis to measure the impact of object properties such as color, texture, and real-world size on detection performance. We replicate, simplify and extend this work on the larger COCO dataset and on image transformations.
Russakovsky~\etal~\cite{russakovsky2013detecting} analyzed the ImageNet localization task and emphasized on fine-grained recognition. Zhang~\etal~\cite{zhang2016far} measured how far we are from solving pedestrian detection. 
Vondrick~\etal~\cite{vondrick2013hoggles} proposed a method for visualizing object detection features to gain insights into their functioning. Some other related works in this line include~\cite{li2019analysis,zhu2012we,zhang2014predicting}. 


The second line regards research in {\bf comparing object detection models}. Some works have analyzed and reported statistics and performances over benchmark datasets such PASCAL VOC~\cite{everingham2010pascal,everingham2015pascal}, MSCOCO~\cite{lin2014microsoft}, CityScapes~\cite{cordts2016cityscapes}, and open images ~\cite{kuznetsova2018open}.
Recently, Huang~\etal~\cite{huang2017speed} performed a speed/accuracy trade-off analysis of modern object detectors. 
Dollar~\etal~\cite{dollar2011pedestrian} and Borji~\etal~\cite{borji2015salient,borji2012quantitative,borji2012state} compared models for person detection, and salient object detection, respectively. 
In \cite{michaelis2019benchmarking}, Michaelis ~\etal assessed detection models on degraded images and observed about 30–60\% performance drop, which could be mitigated by data augmentation. In order to resolve the shortcomings of the AP score, some works have attempted to introduce alternative~\cite{hall2018probability} or complementary evaluation measures~\cite{oksuz2018localization,rezatofighi2019generalized}. A large number of works have also assessed object recognition models and their robustness (\eg~\cite{russakovsky2015imagenet,azulay2018deep,recht2019imagenet,mishkin2017systematic}).

Works in the third line study the {\bf role of context in object detection and recognition} (\eg~\cite{bar2004visual,wolf2006critical,marat2012influence,heitz2008learning,torralba2001statistical,rabinovich2007objects,rosenfeld2018elephant,galleguillos2010context}. Heitz~\etal~\cite{heitz2008learning} proposed a probabilistic framework to capture contextual information between “stuff” and “things” to improve detection. Barnea~\etal~\cite{barnea2019exploring} utilized co-occurrence relations among objects to improve the detection scores. Divvala~\etal~\cite{divvala2009empirical} explored different types of context in recognition. See also~\cite{heitz2008learning,chen2018context,song2011contextualizing,hu2018gather,marat2012influence,alamri2019contextual}. 

\section{Experimental Setup}
\vspace{-1pt}

\vspace{-4pt}
\subsection{Benchmarks}
\vspace{-4pt}
We base our analysis on two recent large-scale object detection benchmarks: \emph{MMDetection}~\cite{mmdetectron,chen2019mmdetection} and \emph{Detectron2}~\cite{Detecron2}. The former evaluates more than 25 models. The latter includes several variants of FastRCNN~\cite{girshick2015fast}. In both benchmarks, all COCO models have been trained on \emph{train2017} and evaluated on \emph{val2017}. Here, we use \emph{MMDetection} to train and test additional models on a new dataset.


\vspace{-1pt}
\subsection{Models}
\vspace{-4pt}
We consider the latest models published in the major vision conferences and the ones included in the above benchmarks. Several variants of the RCNN model including FasterRCNN~\cite{ren2015faster}, MaskRCNN~\cite{he2017mask}, RetinaNet~\cite{lin2017focal}, GridRCNN~\cite{lu2019grid}, LibraRCNN~\cite{pang2019libra}, CascadeRCNN~\cite{cai2018cascade}, MaskScoringRCNN~\cite{huang2019mask}, GAFasterRCNN~\cite{zhu2019empirical}, and Hybrid Task Cascade~\cite{chen2019hybrid} are considered. We also include SSD~\cite{liu2016ssd}, FCOS~\cite{tian2019fcos}, and CenterNet~\cite{zhou2019objects}. Different backbones for each model are also taken into account. 

\subsection{Datasets}
\vspace{-4pt}
We employ 4 datasets including PASCAL VOC~\cite{everingham2015pascal}, our home-brewed FASHION dataset, MSCOCO~\cite{lin2014microsoft}, and OpenImages~\cite{kuznetsova2018open}. Over VOC, we use \emph{trainval0712} for training (16,551 images, 47,223 boxes) and \emph{test2007} (4,952 images, 14,976 boxes) for testing. This dataset has 20 categories. Our FASHION dataset covers 40 categories of clothing items (39 + humans). Trainval, and test sets for this dataset contain 206,530 images (776,172 boxes) and 51,650 images (193,689 boxes), respectively. Fig.~\ref{fig:fashion} displays samples from this dataset (see Supplement for stats). This is a challenging dataset since clothing items are non-rigid as opposed to COCO or VOC objects. MSCOCO has 80 categories. It has carried the torch for benchmarking advances in object detection for the past 6 years. We use \emph{train2017} for training (118,287 images, 860,001 boxes) and \emph{val2017} (5,000 images, 36,781 boxes) for testing. Finally, we use the OpenImages V4 dataset, used in Kaggle competition~\cite{kaggle}. It has 500 classes and contains 1,743,042 images (12,195,144 boxes) for training and 41,620 images (226,811 boxes) for validation (used here for testing).  

\subsection{Metrics}
\vspace{-4pt}
We use COCO API~\cite{cocoeval} to measure AP over IOU thresholds of 0.5 and 0.75 as well as the average AP over IOUs in the range 0.5:.05:0.95. APs are calculated per class and are then averaged. We also report breakdown APs over small (area$\leq 32^2$), medium ($32^2<$area$\leq 96^2$), and large (area$>96^2$) objects. Please see~\cite{cocoeval,palida,Kemal} for details.  

\section{Characterizing the Empirical Upper Bound}
\vspace{-2pt}
We hypothesize that the empirical upper bound in AP is the score of a detector with ground truth bounding boxes labeled by the best object classifier. The classification score is considered as the detection score. This way we essentially assume that the localization problem is solved and what remains is only object recognition. 
However, it might be possible to improve upon this detector in at least two ways: a) by exploiting local context around an object to improve classification accuracy and hence better UAP, and b) by searching over the scene and finding boxes that are easier to classify (compared to the target box) and have enough overlap with the target box. This does not matter for the perfect IOU but may affect IOUs lower than one. 
We carefully investigate these possibilities in the following.

\begin{figure}[t]
\begin{center}
  \includegraphics[width=\linewidth]{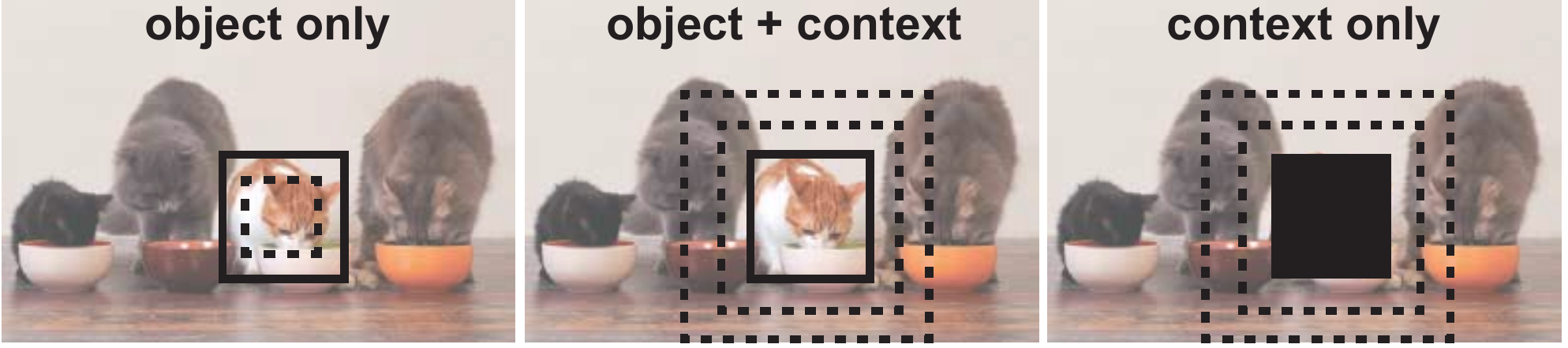}
\end{center}
\vspace{-10pt}
  \caption{Illustration of visual context surrounding an object.}
  \vspace{-5pt}
\label{fig:context}
\end{figure}

\begin{table}
\begin{center}
\begin{scriptsize}
\renewcommand{\tabcolsep}{1pt}

\begin{tabular}{l|ccccc|ccccc|ccc}
\multirow{2}{*}{\textbf {Dataset}} & \multicolumn{5}{c|} {\textbf{object only}} & \multicolumn{5}{c|} {\textbf{object + context}} & \multicolumn{3}{c} {\textbf{context only}} \\
 \cline{2-14} 
 & 0.2 &  0.4 & 0.6 & 0.8 & 1 & 1.2 & 1.4 & 1.6 & 1.8 & 2 &      1.2 & 2 & all img  \\
\hline\hline
VOC & 39.3 &	68.0	& 82.6&	92.5	& {\bf 94.8}&	93.0 &	91.6&	90.6&	88.6	& 87.0   & 63.6 &	64.9	& 35.3     \\
FASHION & - &	52.9	& 66.4	&71.7	& {\bf 88.8}	&82.3&	77.2 &	71.8  & 67.9 &		64.8  &    29.0	& 32.2	&	12.0\\
COCO & - & - & 67.1	 & 79.8 & 	{\bf 86.7} & 82.9	 & 78.3 & 	72.5	 & 67.4 & 	63.0    & 43.7 &	48.9	&	11.0\\
OpenImg. & - & - & -& - & {\bf 69.0} & 65.1 & 62.7 & - & -& - & - &  - & - \\

\end{tabular}
\end{scriptsize}
\end{center}
\vspace{-2pt}
\caption{Recognition accuracy using object and/or its context.}
\vspace{-15pt}
\label{tab:context}
\end{table}

\subsection{Utility of the surrounding context}
\vspace{-4pt}
We trained ResNet152~\cite{he2016deep} (see supp.) on target boxes in three settings as shown in Fig.~\ref{fig:context}: I) \emph{object only}, II) \emph{object + context}, and III) \emph{context only}. Standard data augmentation techniques including mean pixel subtraction, color jittering, random horizontal flip and random rotation (10 degrees) were applied. Boxes were resized to 224 $\times$ 224 pixels and models were trained for 15 epochs. Trained models were tested on the original object box. Results (top-1 accuracy) shown in Table~\ref{tab:context} reveal that the canonical object size contains the most information regarding the category of an object over all four datasets. Increasing or decreasing object box lowers the performance. Context-only scenario leads to high classification score but still does below other cases. Stretching the context to the whole scene drops the performance significantly. 
Training and testing models on the same condition (\ie both on \emph{object+context}) results in higher accuracy on that specific condition but does not lead to better overall recognition accuracy. 

\subsection{Searching for the best label}
\vspace{-4pt}
Essentially the problem definition here is how we can get the best classification accuracy for recognition of objects in the scene by utilizing all the information in the scene. This is different than recognition approaches that treat objects in isolation. Note that recognition accuracy is not the same as AP, since detection scores also matter in AP calculation.

Having the best classifier at hand, we are ready to approximate the empirical upper bound in AP. Before delving into details lets recap how AP is calculated.

\noindent {\bf AP calculation}. For each category, detections over all images are sorted according to their confidences. Starting from the top of this list, the target with the highest IOU with each detection is considered. We have a true positive (TP; hit) if their IOU is $\geq thresh$, and if that target has not been assigned yet. We have a false positive (FP) if IOU$<thresh$ (\ie localization error) or if the target has been assigned (\ie duplicate; two predictions on the same target). A target box can be matched with only one detection (the one with the highest confidence score and IOU$\geq thresh$). If a detection has IOU$\geq thresh$ with two targets, it is assigned to the one with the highest IOU which is not assigned already. Scanning the sorted detection list again, a precision for each recall is obtained and is used to draw the Recall-Precision (RP) curve and to compute the AP. See~\cite{palida,Kemal} for details.

\begin{figure}[t]
\begin{center}
   \includegraphics[width=\linewidth]{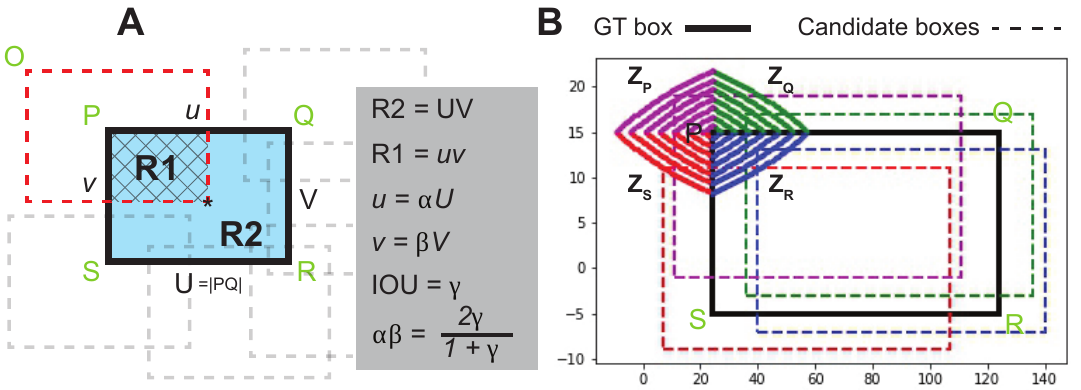}
\end{center}
\vspace{-9pt}
   \caption{A) Illustration of our setup for finding boxes with IOU $\geq \gamma$ with the target box (corresponding to $\alpha\beta = 2\gamma/(1+\gamma)$; $\alpha\beta=2/3$ for $IOU=0.5$), B) The solutions are 4 curves represented by Eqs. 4 to 8. Four sample rectangles are shown with dashed lines.}
\label{fig:synthetic_rects}
\vspace{-17pt}
\end{figure}

We explore two strategies in pursuit of the upper bound AP. In the {\bf first strategy}, we apply the best classifier from the previous section to the target boxes. The detector built in this fashion gives the same AP regardless of the IOU threshold, since our detections are target boxes. As we argued above, it is not possible to improve this detector at IOU=1. However, if we are interested in upper bound for a lower IOU (say $\gamma$), then it might be possible to do better by searching among the candidate boxes near a target box and choose the one that can be classified better than the target box, or aggregate information from nearby boxes. Thus, in our {\bf second strategy}, we sample boxes around an object and either apply the original classifier (trained on canonical object size) or train and test new classifiers on the surrounding boxes. In any case, we always keep the target box but change its label and/or its confidence. First, lets take a look at our box sampling approach, which is illustrated in Fig.~\ref{fig:synthetic_rects}.

\noindent {\bf Sampling boxes with IOU above a threshold.} 
Here, we are interested in finding the coordinates of the top-left corner of all rectangles with IOU $\geq \gamma \ \ \big(\gamma \leq 1\big)$ with the ground-truth bounding box. We use the coordinate system centered at the top-left corner of the target box $P$; which can be easily converted to the image level coordinate frame. The bottom-right coordinates of the desired rectangles that intersect with the target box from the top-left follow the equation $\alpha \beta = 2\gamma/(1 + \gamma)$, where $\alpha$ and $\beta$ are width and height of rectangles, respectively (we assume all boxes have the same width and height as the target box). According to the illustration in Fig.~\ref{fig:synthetic_rects}(A), we have:
\begin{equation}
\text{R1 = uv}  \text{, \ \  } \text{R2 = UV}  \text{, \ \  } \text{IOU} = \gamma  \text{, \ \  } \text{IOU} = \frac{R1}{2R2 - R1}   
\end{equation}
From these equations and assuming $u = \alpha U$, and $v = \beta V$, it is easy to derive the following equations:
\begin{equation}
    R1 = \alpha U \beta V \text{, \ \  } R1 = \frac{2\gamma}{1 + \gamma} R2  \\
    \vspace{-5pt}
\end{equation}
\vspace{-5pt}
and also:
\begin{equation}
    \alpha \beta = \frac{2\gamma}{1 + \gamma} \text{, \ \ }  \alpha \beta = \frac{2}{3} \text{\ \ for \ \ }  \gamma = 0.5 
\end{equation}
The same equation governs the coordinates of the bottom-left, top-left, and top-right corners of the rectangles intersecting with the target box at points $Q$, $R$, and $S$, respectively (in the coordinate frames centered as these points, in order). Calculating the top-left corner of these rectangles (in their corresponding coordinate frames) and representing them in the coordinate frame of point $P$, we arrive at the following four equations (note that these are not lines): 
\begin{eqnarray}
Z_P : \ \ \big\langle~~ (\alpha - 1) U + x_P, \ \ (\beta - 1) V + y_P  ~~\big\rangle \\
Z_Q : \ \ \big\langle~~ (1 - \alpha) U + x_P, \ \ (\beta - 1) V + y_P ~~\big\rangle \\
Z_R : \ \ \big\langle~~ (1 - \alpha) U + x_P, \ \ (1 - \beta) V + y_P ~~\big\rangle \\
Z_S : \ \ \big\langle~~ (\alpha - 1) U + x_P, \ \ (1 - \beta) V + y_P ~~\big\rangle \\
\forall \text{ \ } \alpha,\beta \leq 1, \text{\ \ s.t. \ \ } \alpha \beta = \frac{2\gamma}{1 + \gamma}
\end{eqnarray}

\begin{table}[t]
\begin{center}
\begin{scriptsize}
\renewcommand{\tabcolsep}{3.7pt}

\begin{tabular}{l|c|cccc|cccc} 


 \multirow{2}{*}{\textbf {Dataset}} & \multirow{2}{*}{\textbf {Acc.}} &  \multicolumn{4}{c|} {\textbf{Most Confident Box}} &  \multicolumn{4}{c} {\textbf{Most Frequent Label}} \\ 
 
\cline{3-10}

 & & $AP$ & $AP_{l}$ & $AP_{m}$ &  $AP_{s}$ & $AP$ & $AP_{l}$ & $AP_{m}$ &  $AP_{s}$  \\ 
 
\hline\hline
VOC & 93.7 & 88.7 & 91.7 & 81.4 & 63.8 &  89.1 & 92.0 & 82.9 & 60 \\
FASHION &   87.4 & 68.1 & 68.6 & \underline{61.9} & \underline{49.5}  & 67.7 & 68.2 & \underline{60.7} & \underline{47.8} \\
COCO & 84.8 & 76.9 & 81.8 & 80.6 & \underline{62.8} & 76.4 & 82.0 & 80.4 & 60.7 \\

\end{tabular}
\end{scriptsize}
\end{center}
\vspace{-3pt}
\caption{Results of our second strategy for estimating the upper bound AP (\ie searching for the best bounding box or object label near a target box; among boxes with IOU $\geq 0.5$). Notice that upper bound for AP, AP$0.5$ and AP$0.75$ are all the same. Underlined numbers show where we could improve over the 1st strategy.}
\label{tab:control_res}
\vspace{-10pt}
\end{table}

Using the above equations, we then sample some (here = 4) rectangles with $IOU\geq \gamma$ (Fig.~\ref{fig:synthetic_rects}(B)) and label them with the label of the target box. We then train a new classifier (same ResNet152 as above) on these boxes. This is effectively a new data augmentation technique. Notice that AP is a direct consequence of the classification accuracy, so if we can better classify objects we can obtain a better AP. To estimate UAP, we 
sample a number of rectangles (=4) near a target box (all with $IOU\geq \gamma$), and then label the target box with: a) the label (and confidence) of the box with the highest classification score (\ie most confident box), or b) the most frequent label among the nearby boxes (with the maximum confidence score among them).

\subsection{Upper bound results}
\vspace{-4pt}
Here, we report classification scores, upper bound APs, score of the models (mean AP over all IOUs; unless specified otherwise), and the breakdown AP over categories. 

\noindent {\bf Comparison of strategies.} Summary results of the first strategy are shown in Fig.~\ref{fig:coco_fashion_voc}. As expected UAPs over all IOUs are the same and are much better than the models. To our surprise, our second strategy did not lead to better UAP values, except for few cases including UAPs over medium and small objects on FASHION dataset and small objects on COCO (using most confident boxes), as shown in Table~\ref{tab:control_res}. Applying the original classifier, instead of training new ones on surrounding boxes, or only sampling boxes with higher IOU (\eg 0.9) did not improve the results. Also, setting the confidence of detections to 1 lowers the UAP.
We attribute the failure of the 2nd strategy to the fact that the surrounding boxes may contain additional visual content which may introduce noise in the labels. This leads to a lower classification accuracy and hence a lower AP. Therefore, in what follows we only discuss the results from the first strategy.

\begin{figure}[t]
\begin{center}
\vspace{-5pt}
   \includegraphics[width=\linewidth]{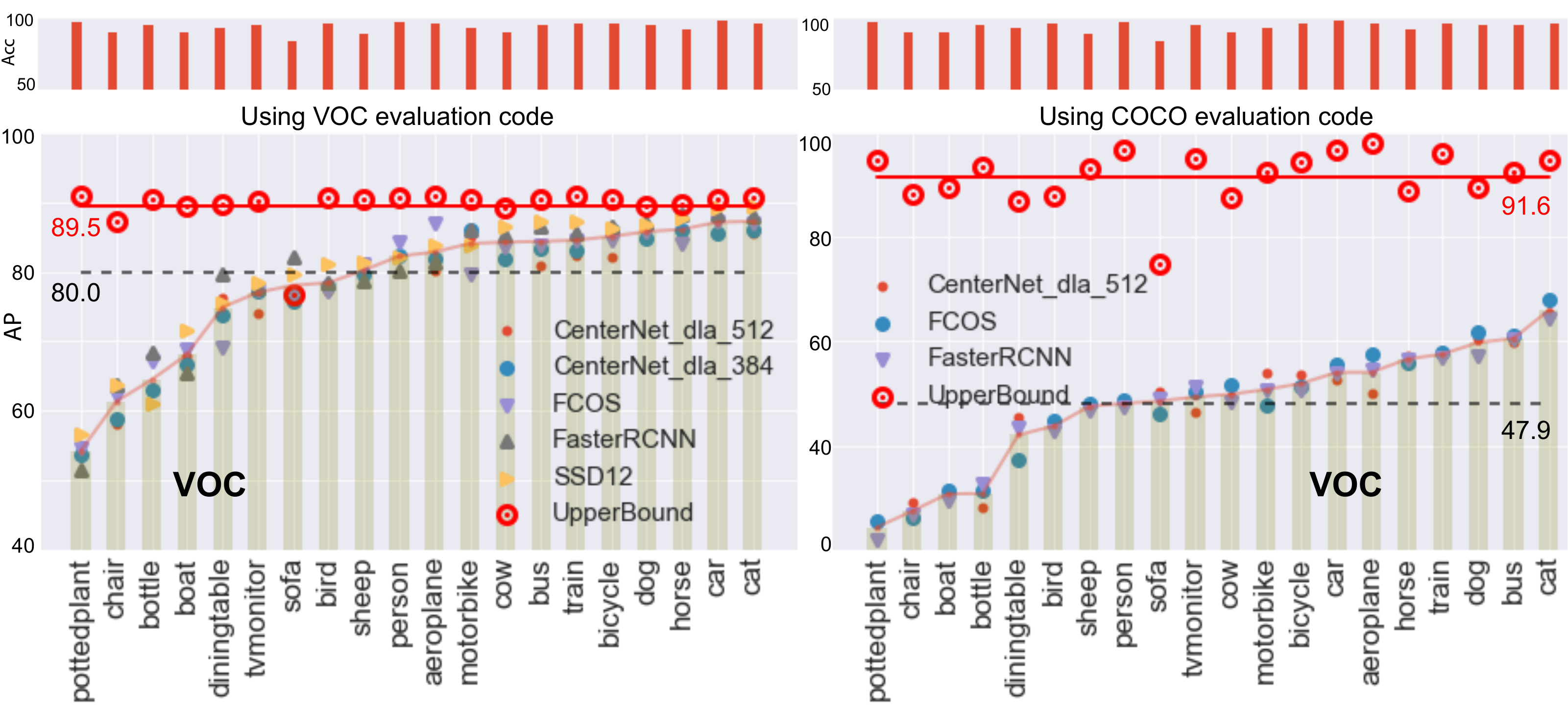}
\end{center}
\vspace{-10pt}
   \caption{Model scores and upper bound AP over PASCAL VOC dataset using VOC (left) and COCO APIs (right). Categories are sorted based on the average model AP. Bar charts show classification scores. Solid red and dashed black lines represent upper bound AP, and the best model AP, respectively.}
\vspace{-3pt}
\label{fig:voc}
\end{figure}

\begin{figure}[t]
\begin{center}
\vspace{-3pt}
   \includegraphics[width=\linewidth]{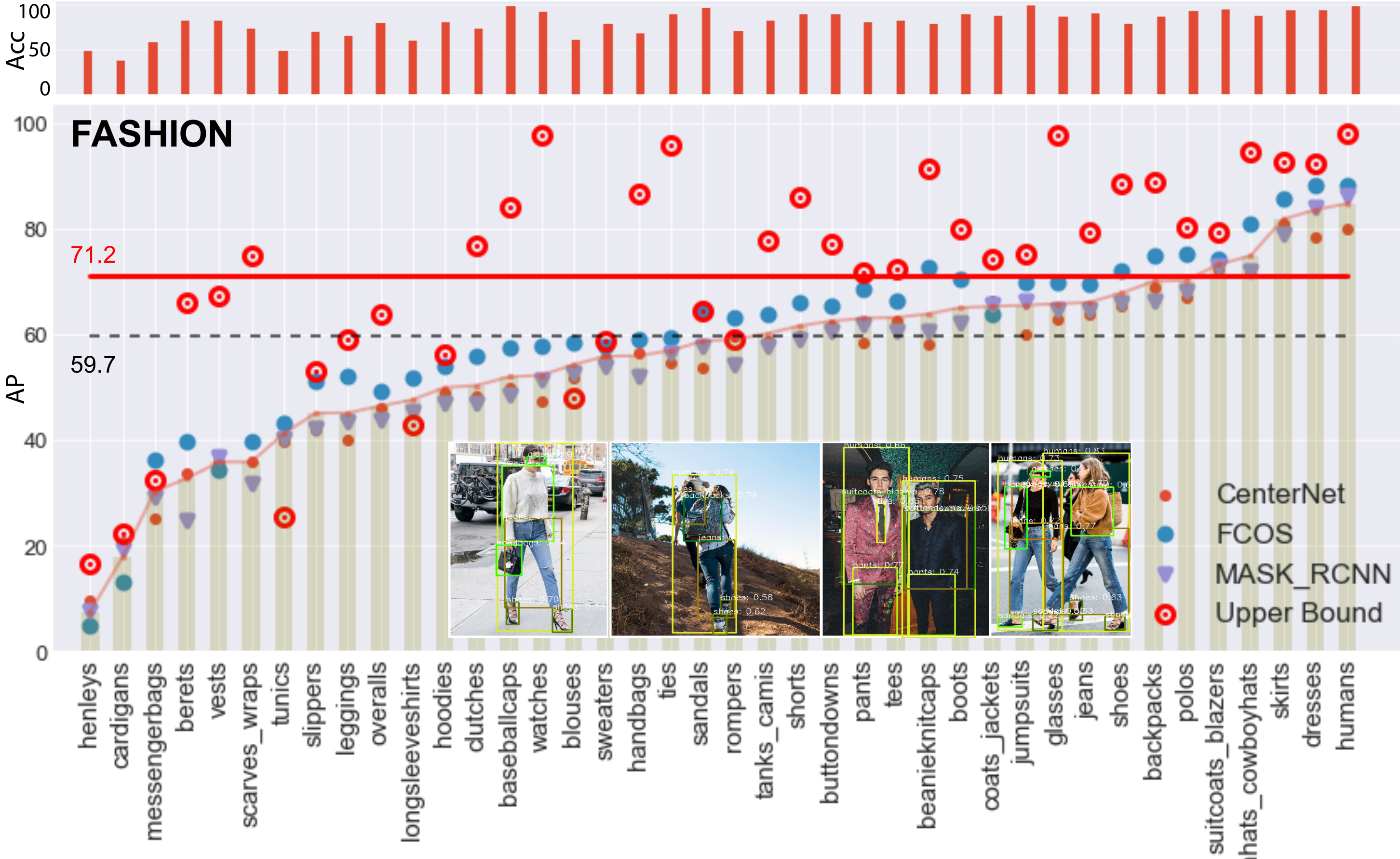}
\end{center}
\vspace{-10pt}
   \caption{upper bound and model APs over the Fashion dataset.}
\label{fig:fashion}
\vspace{-12pt}
\end{figure}

\begin{figure*}[t]
\begin{center}
\vspace{-6pt}
   \includegraphics[width=17.7cm, height=6.4cm]{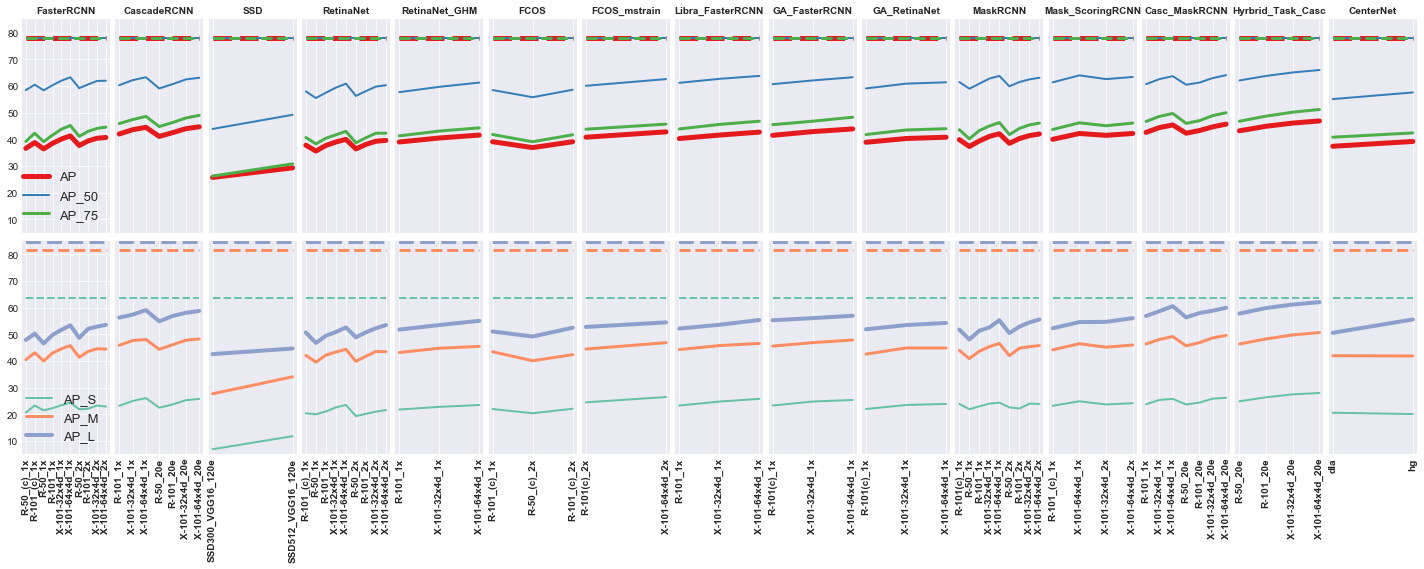}
\end{center}
\vspace{-12pt}
   \caption{APs over COCO dataset borrowed from the \emph{MMDetection} benchmark~\cite{mmdetectron}. We add CenterNet results to \emph{MMDetection}.}
\label{fig:coco_mmdetectron}
\vspace{-12pt}
\end{figure*}

\noindent {\bf PASCAL VOC}. Fig.~\ref{fig:voc} shows results using both VOC and COCO evaluation APIs. The VOC evaluation code is based on IOU=0.5 and calculates the area under the PR curve slightly different than COCO. For VOC, we adopt the code from the CenterNet repository~\cite{CenterNetGitHub}. We have trained and tested 5 models on this dataset including FasterRCNN, FCOS, SSD512, and two variants of CenterNet. The classification accuracy on VOC is very high (94.7\%). Consequently, the UAP is very high (91.6 using the COCO API). FCOS model does the best here with AP of 47.9 (right  panel in Fig.~\ref{fig:voc}; dashed lines). As it can be seen, there is a large gap between the AP of the best model and the UAP on this dataset ($\sim$45). Models are consistent in their performance across different categories.

\noindent {\bf FASHION}. Results are shown in Fig.~\ref{fig:fashion}. The best classification accuracy on this dataset is 88.8\% (Table~\ref{tab:context}, and supplement). The UAP is 71.2 and the AP of the best model is 59.7 (FCOS). Interestingly, FCOS performs quite close to the upper bound at IOU=0.5 (Fig.~\ref{fig:coco_fashion_voc}). Models perform better here than over VOC. The FASHION UAP is lower than VOC UAP perhaps because classification is more challenging on the former dataset. The gap between UAP and model AP here, however, is much smaller than VOC. This could be partly due to the fact that FASHION scenes have less clutter and larger objects than the VOC scenes. While per-class UAP is above the AP of the best model over all VOC classes, UAPs of 5 FASHION categories fall below the best model AP (\emph{messenger bags, tunics, long sleeve shirts, blouses, and rompers}). Looking at the classification scores, we find that they have a low accuracy.

\noindent {\bf COCO}. Existing benchmarks have provided an efficient ecosystem for developing, evaluating and comparing detection models especially on the COCO dataset. They provide trained models over a variety of settings. 
Borrowing the \emph{MMDetection} benchmark and adding the results from CenterNet to it, we end up comparing 15 models (71 in total; combination of models and backbones). Model scores are shown in Fig.~\ref{fig:coco_mmdetectron}. The best models here are Hybrid Task Cascade model~\cite{chen2019hybrid} and Cascade MaskRCNN~\cite{cai2018cascade}, with APs of 46.9 and 45.7, respectively. See supplement for Recall results. 
The upper bound AP on COCO is about 78.2. Recall that UAP does not depend on the IOU threshold since detected boxes are classified ground truth targets. The gap between the best model AP and UAP is above 30. The gap is much smaller for AP at IOU=0.5 which is about 10. The UAP is much lower over small objects than UAP over large objects. This also holds for models. The gap between UAP and model AP over small objects is about 35 which is much higher than the gap over medium or large objects.

\begin{figure*}[t]
\begin{center}
\vspace{-6pt}
   \includegraphics[width=\linewidth]{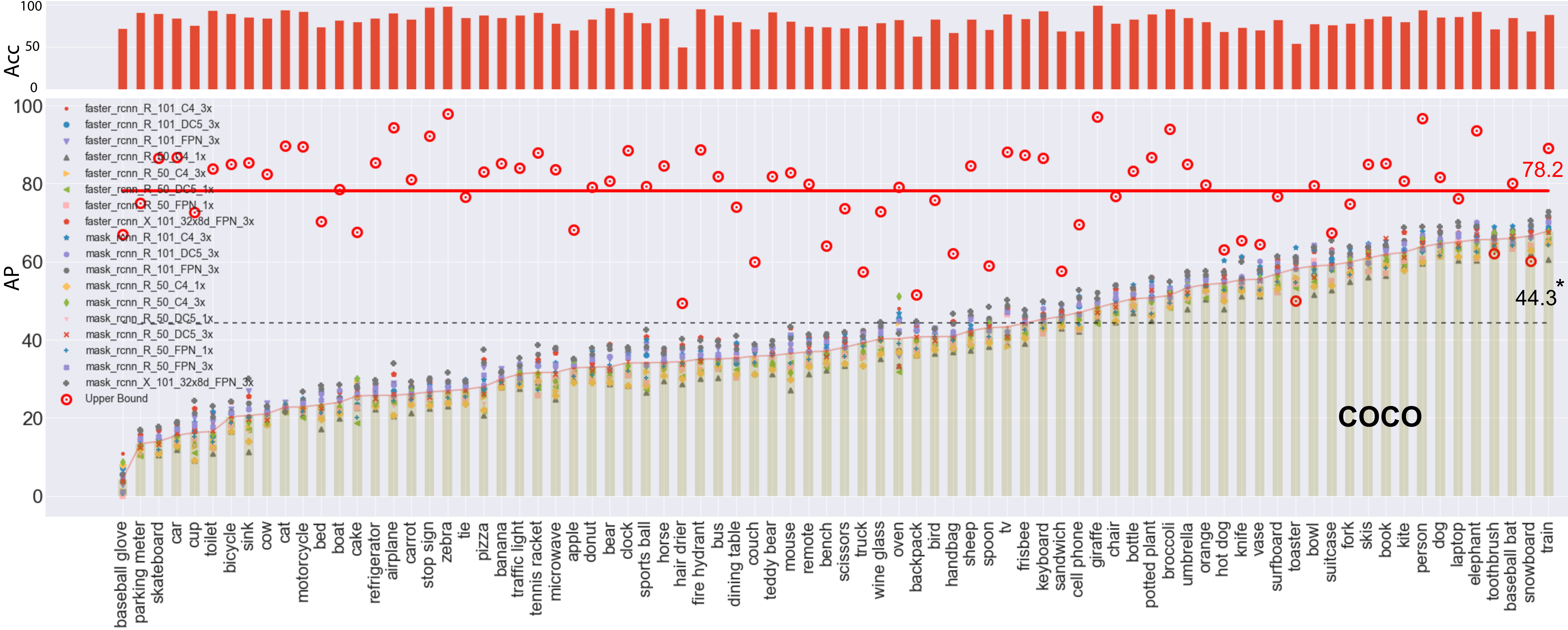}
   
\end{center}
 \vspace{-10pt}
   \caption{Detection APs over MSCOCO dataset borrowed from \emph{Detectron2} benchmark~\cite{Detecron2}. The horizontal dash line corresponds to the best model among the shown models. ``*": The best AP here is 44.3 which is smaller than the best so far on COCO (46.9). See also Fig.~\ref{fig:coco_fashion_voc}.}
\label{fig:coco_detectron}
\vspace{-10pt}
\end{figure*}

Breakdown APs over object categories are shown in Fig.~\ref{fig:coco_detectron}. For this analysis, we use the \emph{Detectron2} benchmark which reports per-category results mainly over the RCNN model family. We noticed that the aggregate scores on \emph{MMDetection} and \emph{Detectron2} are quite consistent.
Among the 18 variants of Faster-RCNN and MASK-RCNN, the best model has the AP of 44.3 (shown by the dashed line) which is lower than the best available model on COCO (46.9; Fig.~\ref{fig:coco_fashion_voc}) and the upper bound AP. Among 80 classes, only three (\emph{snowboard, toothbrush, and toaster}) have UAPs below the best model APs.

\noindent {\bf OpenImages}. This dataset~\cite{kuznetsova2018open} is the latest endeavor in object detection and is much more challenging than its predecessors. 
Our classifier achieves 69.0\% top-1 accuracy on the validation set of OpenImages V4 which is lower than other the three datasets. We achieve 58.9 UAP, using the TensorFlow evaluation API for computing AP~\cite{openImagesLink} on this dataset, which is different than COCO AP calculation (here we discarded grouping and super-category). We are not aware of any model scores on this set of OpenImages V4.

\begin{figure}[t]
\begin{center}
   \includegraphics[width=\linewidth]{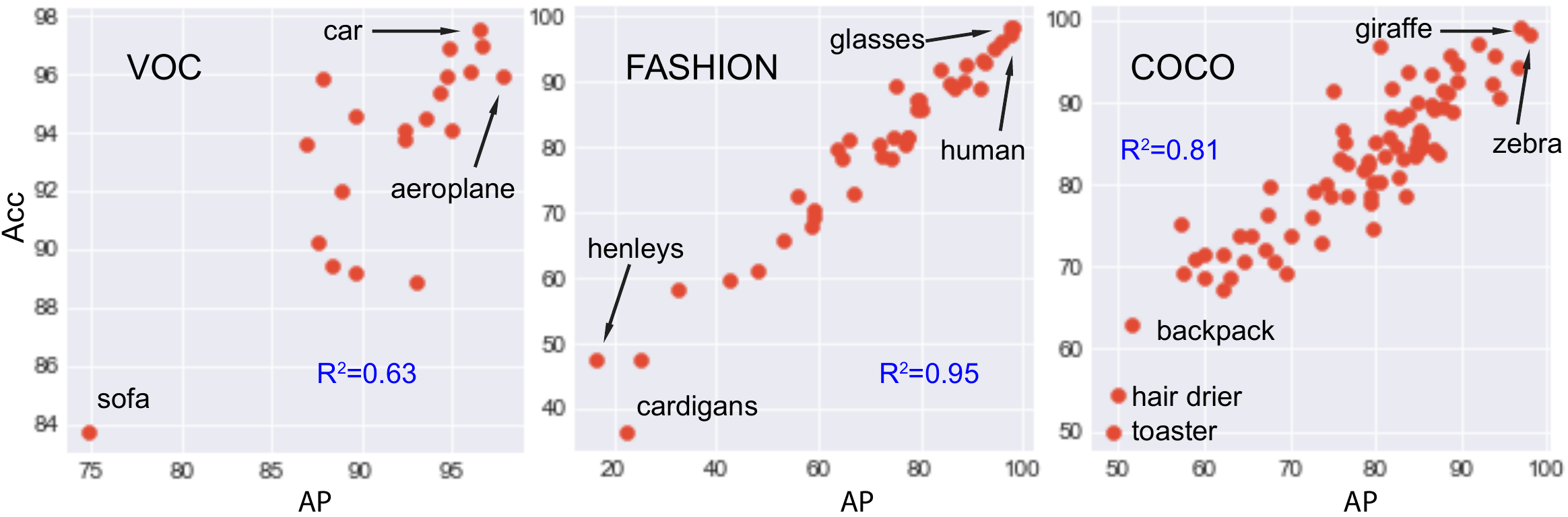}
\end{center}
\vspace{-12pt}
   \caption{Correlation between classification accuracy and upper bound AP. The higher the Acc., the better the UAP.}
 \vspace{-15pt}
\label{fig:AP_ACC_correlation}
\end{figure}

\noindent {\bf AP vs. classification accuracy}. We found that there is a linear positive correlation ($R^2$ = 0.81 on COCO) between the UAP and the classification accuracy (Fig.~\ref{fig:AP_ACC_correlation}). The higher the classification accuracy, the higher the UAP. We did not find a correlation between the accuracy and model APs, nor between the object size and accuracy (or UAP). The dependency of UAP on accuracy, highlights the importance of recognition on object detection and constitutes the core of our analyses in the next two sections.

\vspace{-6pt}
\section{Error Diagnosis}
\vspace{-6pt}

\begin{figure}[t]
\begin{center}
   \includegraphics[width=\linewidth]{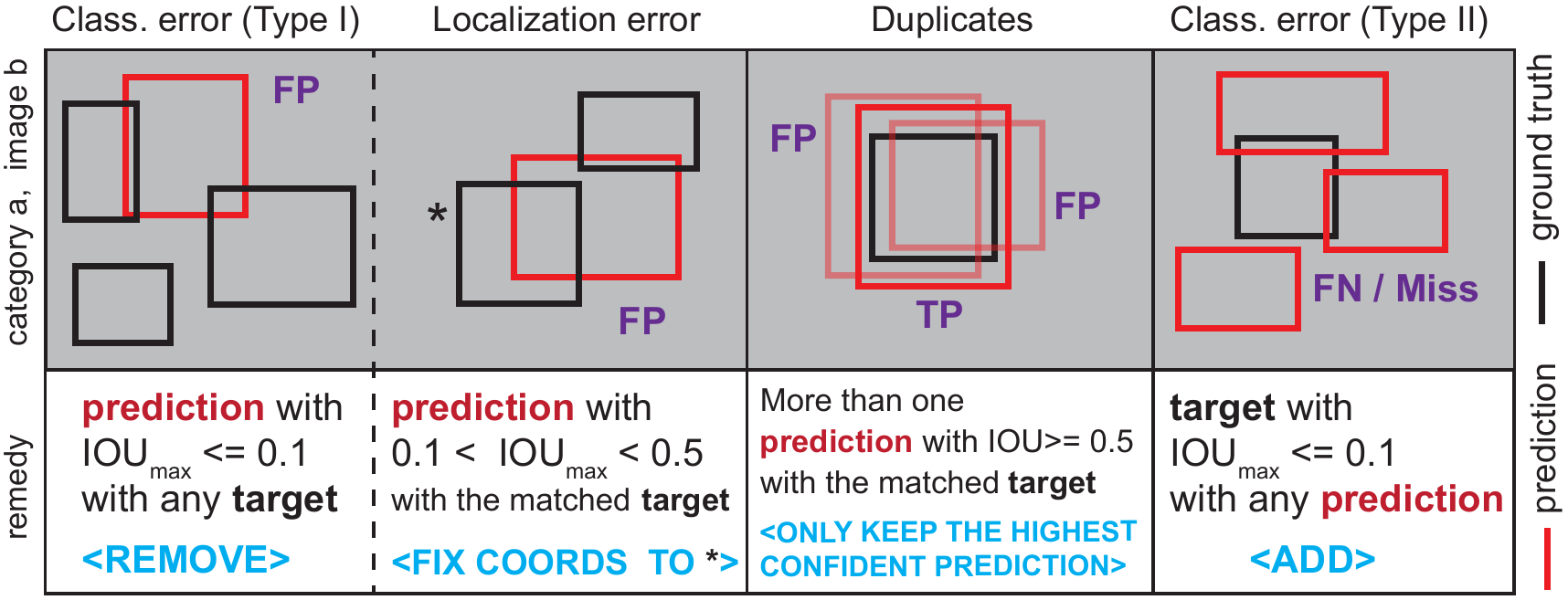}
\end{center}
\vspace{-10pt}
   \caption{Illustration of different error types in object detection.}
\label{fig:error_types_illustration}
\vspace{-3pt}
\end{figure}

To pinpoint the shortcomings of object detectors, we follow the analysis by Hoeim~\etal~\cite{hoiem2012diagnosing}, but revise it in two major ways. First, instead of inspecting errors across categories, here we perform a per-category error analysis (\ie binary manner). This simplifies the process and makes it easier to understand. See Fig.~\ref{fig:error_types_illustration}. We combine all types of class confusions (\eg similar classes, other classes, and background in Hoeim~\cite{hoiem2012diagnosing}) into two types of classification errors: a) \emph{confusion with the background} (Type I), and 2) \emph{misses} (Type II). Notice that this implicitly contains the above misclassification types but is much easier to investigate. In fact, recent object detectors such as FCOS~\cite{tian2019fcos} and CenterNet~\cite{zhou2019objects} also adopt this strategy to classify objects (\ie an object is of a particular class or is not). Second, Hoeim~\etal successively remove errors to reach the AP of 1. We argue that this approach convolutes different error types and does not correctly reflect the true contribution of errors (\ie understating or exaggerating error types). For example, according to the COCO analysis tool~\cite{cocoeval}, any matches to objects with a different class label but in the same supercategory do not count as either a FP or a TP. Also, the COCO tool removes mislocalized predictions. In this case, we argue that correcting the mislocalized predictions is more effective than removing them because it can reveal other sources of weakness in a model. For example, it may lead to generating duplicates which would have been overlooked by removing the detections. 
In contrast, here we explicitly handle the errors by removing, correcting or adding detections when appropriate. Similar to Hoeim~\etal our analysis is also based on IOU=0.5.

We repeat the following procedure for each category-image pair (shown in Fig.~\ref{fig:error_types_illustration}; from left to right). First, we remove the detections with the maximum IOU$_{max}\leq0.1$ with any target (\ie classification error Type I; confusion with the background). Second, we correct the miss-localized predictions with $0.1<\text{IOU}_{max}<0.5$. In this step, coordinates of these boxes are replaced with their matching target box coordinates (which is the target with the max IOU) while their confidence scores and labels are preserved. Third, duplicates (\ie redundant detections) are removed. An unmatched detection is considered duplicate if it falls (\ie has IOU$\geq0.5$) over a target with an already assigned detection (with higher score). See supplement for details. 
Fourth, eventually, misses are treated. A miss is a target with $\text{IOU}_{max}\leq 0.1$ with any unmatched detection, and is added to the list of detections (with score of 1). Before performing this step, we set the coordinates of detections as the coordinates of their matching targets, since we now know which prediction is paired with which target (\ie one to one mapping; no duplicates).

Results of error diagnosis are shown in Table~\ref{tab:error_type_our_results} for 3 models over 3 datasets. We start from the original detection set and progressively measure the impact of fixing each error type in the order explained above and shown in Fig.~\ref{fig:error_types_illustration}. Confusion with the background (and other classes; see above) has the highest contribution to the overall error, across all models. This indicates that models often falsely confuse background clutter or other classes as a particular object category. The second most important error type is misses. Interestingly, localization error weighs more than duplicates and has higher impact on COCO and VOC datasets than the FASHION dataset, possibly because the former two contain a larger number of small objects. Conversely, over the FASHION dataset, duplicates matter more, perhaps because class confusion is higher (\eg confusion in slippers vs. sandals; different types of hats, etc.).
Models behave almost consistently across the three datasets. 

\begin{figure}[t]
\begin{center}
   \includegraphics[width=\linewidth]{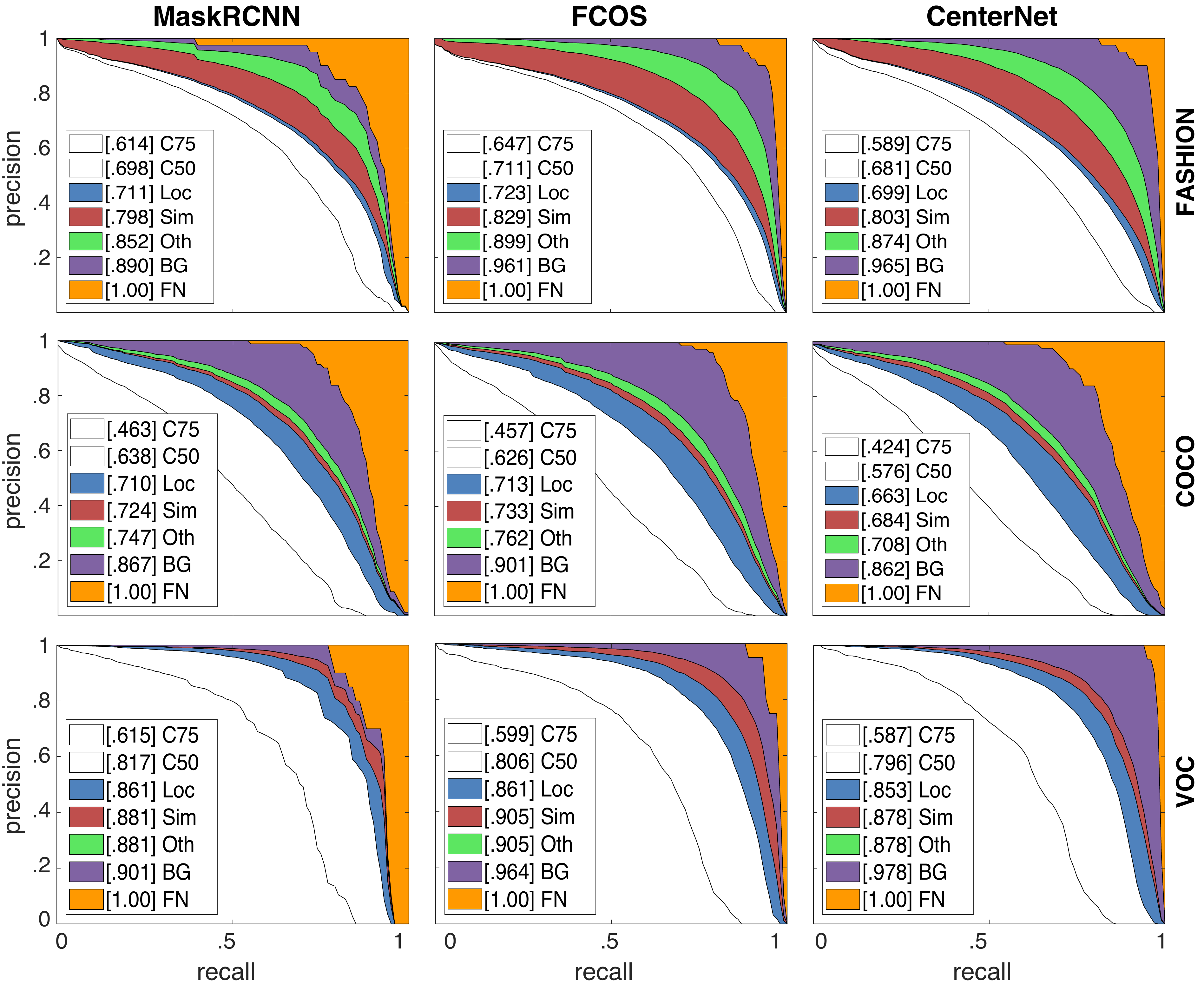}
\end{center}
\vspace{-13pt}
   \caption{Quantifying the contribution of errors in object detection using the COCO analysis code~\cite{cocoeval}.}
\label{fig:error_type_coco_results}
\vspace{-15pt}
\end{figure}

We also cross checked our results with results obtained using the COCO analysis tool (implementing Hoeim~\etal). Notice that numbers from COCO analysis tool are not directly comparable to ours since our strategy is different and, unlike us, it does not explicitly address duplicate errors. Nevertheless, based on APs and PR curves in Fig.~\ref{fig:error_type_coco_results}, we arrive at similar conclusions to ours. Here, again we observe that classification error Type I (Sim, Oth, and BG in Fig.~\ref{fig:error_type_coco_results}) accounts for the largest fraction of errors, followed by misses (FN) and localization (Loc) errors.

\begin{table}
\begin{center}
\begin{scriptsize}
\renewcommand{\tabcolsep}{1.6pt}

\begin{tabular}{l|l|c||c|c|c|c}
{\bf Dataset} & {\bf Model}      & {\bf mAP} & {\bf - Cls. (Type I)} & {\bf + Local.}  & {\bf - Duplicates} & {\bf + Misses} \\   \hline \hline
        & MaskRCNN   & 54.1              & 85.9          & 87.7         & 88.7     & 100   \\ 
FASHION   & CenterNet   & 54.0      & 88.8       & 91.7        & 96.2     & 100           \\ 
        & FCOS        & 59.7              & 90.1     & 91.9    & 95.9     & 100          \\ \Xhline{2\arrayrulewidth}
        & MaskRCNN   & 42.1              & 70.1         & 79.0           & 82.7     & 100     \\  
COCO    & CenterNet   & 39.2      & 66.1      & 78.0            & 81.7     & 100           \\ 
        & FCOS        & 42.8      & 69.6      & 80.8             & 85.4     & 100           \\ \Xhline{2\arrayrulewidth}
        & MaskRCNN & 47.3     & 73.7        & 78.8     & 79.7     & 100           \\ 
VOC  & CenterNet   & 47.8              & 79.0      & 88.5           & 92.6     & 100           \\ 
        & FCOS        & 47.9      & 76.3    & 85.0               & 90.3     & 100           \\          
\end{tabular}
\end{scriptsize}
\end{center}
\vspace{-4pt}
\caption{Quantifying the contribution of errors in object detection. ``Local." and ``Dup." stand for localization error and duplicate removal, respectively. mAP is the model AP over all IOUs. }
\vspace{-13pt}
\label{tab:error_type_our_results}
\end{table}

\vspace{-6pt}
\section{Invariance Analysis}
\begin{figure}[t]
\begin{center}

   \includegraphics[width=\linewidth]{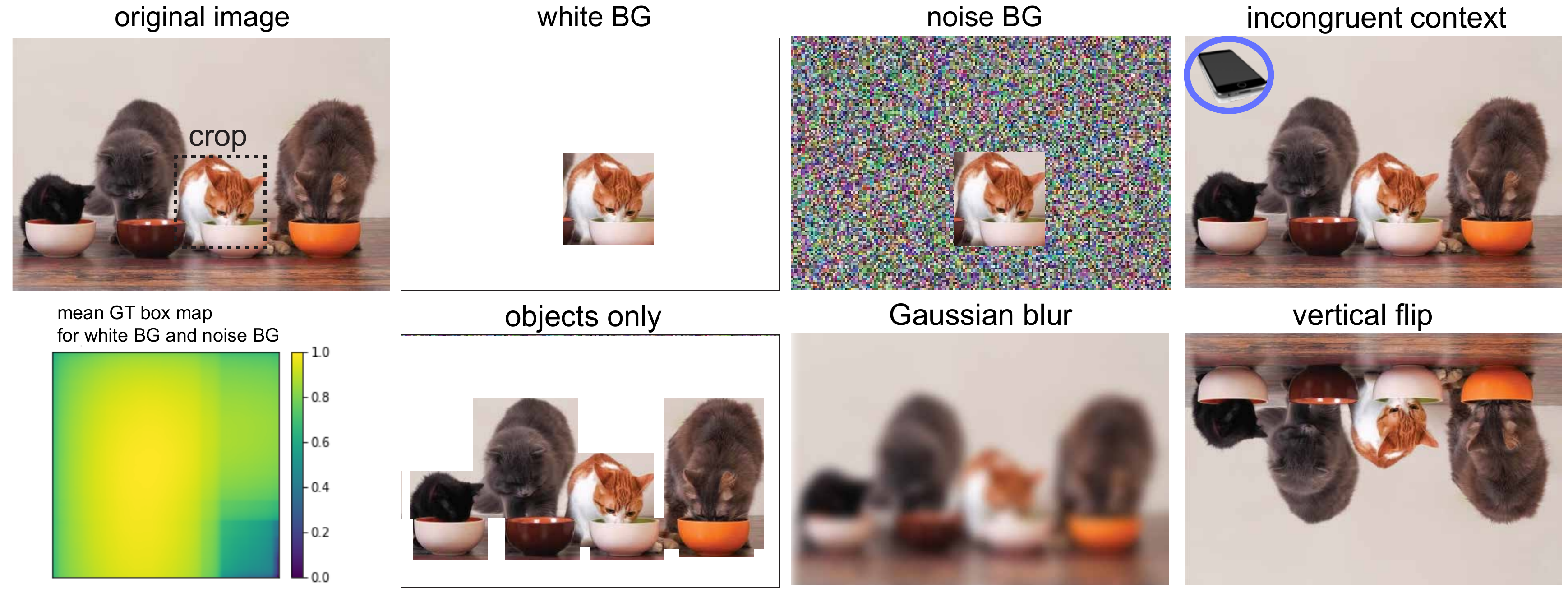}
\end{center}
\vspace{-10pt}
   \caption{Analysis of the impact of context and invariance in object detection. The bottom-left panel shows the distribution of target object boxes (\emph{COCOval2017}) in log scale (See supplement).}
\label{fig:invariance1}
\vspace{-8pt}
\end{figure}

\vspace{-6pt}
Complementary to our error diagnosis, here we conduct a series of experiments to reduce the impact of localization or recognition in detection pipelines (one at a time). Our principal emphasize is on the recognition component. These experiments are performed over the \emph{COCOval2017} set and are illustrated in Fig.~\ref{fig:invariance1}. 
Trained models, over the \emph{COCOtrainval0712} set, are employed.

\begin{figure}[t]
\begin{center}
   \includegraphics[width=\linewidth]{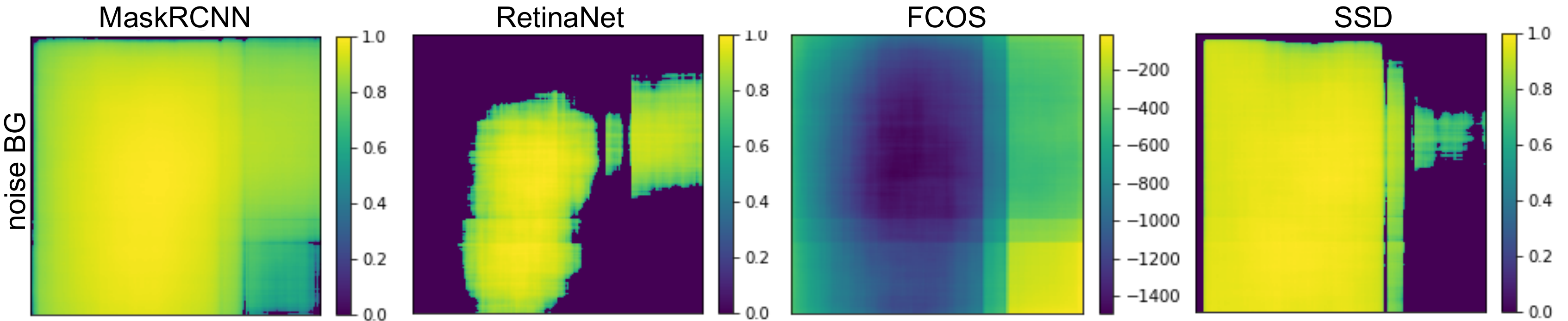}
\end{center}
\vspace{-10pt}
   \caption{\footnotesize{Distribution of predicted boxes on \emph{COCOval2017} (log scale).}}
\label{fig:invariance2}
\vspace{-5pt}
\end{figure}

{\bf Analysis of context.}
In the first experiment, we generated stimuli in which a single object was placed in a white background or in a white noise background (one object per image, hence number of images equal to the number of objects). Contrary to our expectation, 
we found that models either underestimate or overestimate the distribution of target bounding boxes. Fig.~\ref{fig:invariance2} shows the difference in distribution of predicted boxes and distribution of ground-truth boxes. Interestingly, models search all over the place. FasterRCNN and RetinaNet oversample boxes around targets, while FCOS generates a fair amount. This hints towards the shortcomings in objectness prediction in models.
Quantitative results, presented in Table~\ref{tab:invariance_results1}, show that models perform poorly on these images (about the same in both conditions but lower than the original images). They are hindered much more on small objects than medium or large ones, which shows how critical context is for recognition and detection of small objects. 
Interestingly, in white/noise BG and object-only cases, the AP-large increases but the AP-small decreases (compared to orig. images). FCOS, ranking higher on original images, does better here as well. 

In the second experiment, object-only case, we removed the image background and preserved all the objects (hence the same number of images as in \emph{COCOval2017}). To our surprise, FCOS and SSD performed better on these images than the original ones (Column 1 vs. 10 in Table~\ref{tab:invariance_results1}). Compared to the original images, they did better on large objects and lower on small objects in the object-only case.

In the third experiment, we paste objects in incongruent backgrounds (\eg a boat in the street), similar to Rosenfeld~\etal~\cite{rosenfeld2018elephant} but over a larger dataset and including more models (they did not report AP). We paste 9 objects including \emph{bear, keyboard, refrigerator, surfboard, train, tv, cake, horse, and oven} on 100 images taken from the FASHION dataset; 900 images in total. Results are given in Table.~\ref{tab:context_incongruent_results}. Interestingly, models performed well on this dataset. They failed drastically on \emph{surfboard} and \emph{oven} which seem to be a little hard for humans. \emph{Cake, bear, and horse} were the easiest ones. FCOS did the best among models.

\begin{table}
\begin{center}
\begin{scriptsize}
\renewcommand{\tabcolsep}{1.5pt}

\begin{tabular}{l|ccc|ccc|ccc}
\multirow{2}{*}{\textbf {Model}} & \multicolumn{3}{c|} {\textbf{white BG}} & 
\multicolumn{3}{c|} {\textbf{noise BG}}& 
\multicolumn{3}{c} {\textbf{objects\_only}} \\
 \cline{2-10} 
& $AP$ & $AP^{.5}$ & $AP^{.75}$ & $AP$ & $AP^{.5}$ & $AP^{.75}$ & $AP$ & $AP^{.5}$ & $AP^{.75}$ \\
\hline\hline
FasterRCNN & 31.1 & 42.0 & 36.1 & 31.8 & 39.8 & 36.8 & 35.9 & 55.8 & 39.5 \\
RetinaNet & 33.1 & 41.0 & 37.3 & 32.7 & 39.1 & 36.6 & 39.8 & 58.4 & 43.4 \\
FCOS & 34.5 & 42.0 & 37.1 & 34.2 & 39.8 & 37.4 &   43.6 & 60.6 & 46.9 \\
SSD512 & 27.4 & 36.7 & 32.3 & 26.0 & 33.4 & 34 &  30.5 & 48.6 & 32.9 \\
\Xhline{2\arrayrulewidth}
\textbf {Model} & $AP_s$ & $AP_m$ & $AP_l$ & $AP_s$ & $AP_m$ & $AP_l$ & $AP_s$ & $AP_m$ & $AP_l$ \\
\hline
FasterRCNN & 7.5 & 35.9 & 49.9 & 7.0 & 36.6 & 52.1 & 17.5 & 40.6 & 48.6 \\
RetinaNet & 8.3 & 37.5 & 53.2 & 6.4 & 38.3 & 54.2 & 18.9 & 44.5 & 56.4 \\
FCOS & 8.5 & 39.8 & 55.2 & 9.4 & 39.5 & 54.8 & 22.1 & 48.8 & 58.7  \\
SSD512 & 7.0 & 31.4 & 45.1 & 4.6 & 29.3 & 45.2 &  9.8 & 35.7 & 48.4 \\

\end{tabular}
\end{scriptsize}
\end{center}
\vspace{-8pt}
\caption{Results of invariance analysis over \emph{COCOval2017}.}
\vspace{-10pt}
\label{tab:invariance_results1}
\end{table}

\begin{table}
\begin{center}
\begin{scriptsize}
\renewcommand{\tabcolsep}{2pt}

\begin{tabular}{l|ccccccccc|c}
Model & train & horse & bear & surfboard & cake & tv & keyboard & oven & fridge & {\bf Avg.} \\

\hline
\hline

Fa.RCNN & 64.0 & 58.4 & 84.7 & 2.4 & 77.9 & 74.3 & 54.7 & 15.5 & 20.3 & 50.2 \\
RetinaNet  & 54.2& 89.2& 90.6& 2& 85.7& 86.6& 10.1& 24.8& 69.3 & 57.0\\
FCOS       & 73.4& 91.5& 94.0 & 17.1& 87.6& 92.1& 9.8& 44.2& 76.2 & {\bf 65.1}\\
SSD512 .   & 84.3& 58.9& 78.5& 3.8& 76.9& 69.8& 42.6& 8.4& 47.6 & 52.3\\ 
\Xhline{2\arrayrulewidth}
{\bf Avg.} & 69.0& 74.5  & {\bf 87.0} & 6.3 & 82.0 & 80.7 & 29.3 & 23.2 & 53.4 & 56.2

\end{tabular}
\end{scriptsize}
\end{center}
\vspace{-8pt}
\caption{Model APs(IOU=.5) over objects in incongruent contexts.}
\vspace{-15pt}
\label{tab:context_incongruent_results}
\end{table}

{\bf Robustness to image transformations.}
In the fourth experiment, we evaluated models on objects that were a) cropped right out of the image, or b) cropped and resized such that their smallest dimension became 300 pixels (while preserving the aspect ratio). Models performed terribly in both cases, with RetinaNet doing better (Table~\ref{tab:invariance_results2}). Poor performance here demonstrates how sensitive models are to object scale and that they lack robustness to object appearance. Visually inspecting the images, we found it very difficult to recognize the cropped objects, especially the small ones.

Fifth and sixth experiments regard testing models on 
Gaussian blur (with a 11 $\times$ 11 kernel) and vertical flip, respectively. Results in Table~\ref{tab:invariance_results2} 
show that both types of transformations dramatically hinder performance with higher impact for vertical flip. We do not have a baseline for human performance on these cases, but a quick browsing shows that it is still possible to detect objects, albeit with more effort. RetinaNet and FCOS outperform other models here. 

\noindent {\bf Analysis of errors.} Here we measure the impact of each error type in three detection tasks including object-only, Gaussian blur and vertical flip. See Table~\ref{tab:error_type_invariance_results} for results. Error types in order of importance include: \emph{misses, localization, misclassification (Type I), and duplicates}, over three tasks. Models miss more objects in vertical flip and Gaussian blur cases compared to the objects-only case. There is less confusion with BG in objects-only case than original images (classification Type I) since there is no background clutter.

\begin{figure}[t]
\begin{center}
   \includegraphics[width=\linewidth]{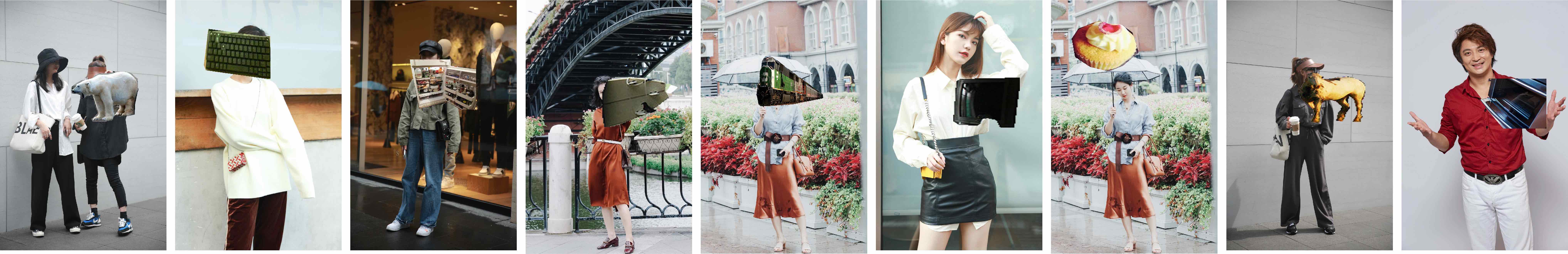}
\end{center}
\vspace{-11pt}
   \caption{\footnotesize{Samples of our dataset of objects in incongruent background.}}
\label{fig:incongruent}
\vspace{-10pt}
\end{figure}

\begin{table}
\begin{center}
\begin{scriptsize}
\renewcommand{\tabcolsep}{.5pt}

\begin{tabular}{l|ccc|ccc|ccc|cc}
\multirow{2}{*}{\textbf {Model}} & \multicolumn{3}{c|} {\textbf{crop}} & 
\multicolumn{3}{c|} {\textbf{Gaussian blur}}& 
\multicolumn{3}{c|} {\textbf{vertical flip}} & \multicolumn{2}{c} {\textbf{orig img.}}\\
 \cline{2-12} 
& $AP$ & $AP^{.5}$ & $AP^{.75}$ & $AP$ & $AP^{.5}$ & $AP^{.75}$ & $AP$ & $AP^{.5}$ & $AP^{.75}$ & $AP$ & $AP^{.5}$ \\
\hline\hline
Fa.RCNN &  8.4 & 15.0 & 8.2 &  17.1 & 29.6 & 17.4 & 15.5 & 27.3 & 15.7 & 36.4 & 58.4\\
RetinaNet & 16.9 & 22.7 & 18.8 &  21.5 & 34.7 & 22.5 & 18.7 & 30.7 & 19.3 & 40.0 & 60.9\\
FCOS & 14.3 & 18.5 & 15.3 &  21.0 & 33.7 & 21.6 & 19.1 & 30.2 & 19.6 & 42.8 & 62.6 \\
SSD512 & 13.4 & 18.9 & 14.9 & 15.1 & 26.6 & 15.2 & 12.1 & 22.2 & 11.9 & 29.3 & 49.2 \\
\Xhline{2\arrayrulewidth}
\textbf {Model} & $AP_s$ & $AP_m$ & $AP_l$ & $AP_s$ & $AP_m$ & $AP_l$ & $AP_s$ & $AP_m$ & $AP_l$ & $AP_s$ & $AP_l$\\
\hline
Fa.RCNN & 0 & 1.3 & 18.7 &     3.8 & 18.3 & 31.5 & 6.2 & 16.6 & 24.7 & 21.5 & 46.6\\
RetinaNet &  1 & 5.2 & 34.1 &    5.1 & 22.8 & 39.0 &  7.5 & 20.5 & 29.5 & 23.5 & 52.6 \\
FCOS & 1 & 4.5 & 32.2 &  5.3 & 22.5 & 37.4 & 8.0 & 20.8 & 30.0 & 26.5 & 54.5 \\
SSD512 & 1 & 2.9 & 25.7 &  2.0 & 15.2 & 30.9 & 4.0 & 12.6 & 22.5 & 11.8 & 44.7 \\

\end{tabular}
\end{scriptsize}
\end{center}
\vspace{-6pt}
\caption{Additional results of invariance analysis over \emph{COCOval2017} dataset. Fa.RCNN = FasterRCNN.}
\vspace{-5pt}
\label{tab:invariance_results2}
\end{table}

\begin{table}
\begin{center}
\begin{scriptsize}
\renewcommand{\tabcolsep}{2pt}

\begin{tabular}{l|l|c||c|c|c|c}
{\bf Dataset} & {\bf Model}      & {\bf mAP} & {\bf - Cls. (Type I)} & {\bf + Local.}  & {\bf - Duplicates} & {\bf + Misses} \\   \hline \hline
objects               & FasterRCNN & 55.8              & 61.5                       & 69.3                               & 75.2     & 100           \\ 
only  & RetinaNet & 58.4              & 64.6                       & 72.6                                & 79.9     & 100           \\ 
               & FCOS       & 60.6              & 67.8                       & 77.0                                & 82.3     & 100           \\ \Xhline{2\arrayrulewidth}
Gaussian               & FasterRCNN  & 29.6              & 37.2                       & 47.4                                & 55.2     & 100           \\ 
 blur          & RetinaNet & 34.7              & 42.3                       & 53.5                                & 64.3     & 100           \\ 
               & FCOS       & 33.7              & 43.1                       & 56.8                                & 65.3     & 100           \\ \Xhline{2\arrayrulewidth}
vertical               & FasterRCNN  & 27.3              & 37.0                       & 49.6                                & 57.3     & 100           \\ 
 flip & RetinaNet & 30.7              & 41.1                       & 54.1                                & 64.6     & 100           \\ 
               & FCOS       & 30.2              & 41.3                       & 57.1                                & 65.6     & 100          \\ 
            
\end{tabular}
\end{scriptsize}
\end{center}
\vspace{-5pt}
\caption{Error analysis of models over transformed images. }
\vspace{-15pt}
\label{tab:error_type_invariance_results}
\end{table}

\vspace{-8pt}    
\section{Conclusion and Outlook}
\vspace{-8pt}
Through exhaustive analyses, we found that a) models perform significantly below what is empirically possible, b) the performance gap is larger over small objects, indicating that scale is one of the major problems in object detection, c) the bottleneck in object detection is object recognition, and d) detection models lack generalization in terms of searching the right places, utilizing context, recognition of small objects, and robustness to image transformation. We did not find a significant contribution from the surrounding context of a target or its nearby overlapping boxes to better classify it. A further investigation of this with extensive data augmentation and optimization may increase the accuracy but is unlikely to drastically improve the UAP. To evaluate the recognition component of a model, one can feed the target boxes to a model and collect its decisions on them. This is, however, cumbersome and needs to be coded for each model separately, whereas our diagnosis tool is general.

We invite researchers to periodically update the upper bound in detection scores including AP and other recently proposed ones such as LIP~\cite{hall2018probability} and probability-based detection quality~\cite{oksuz2018localization}, as new object recognition models surface. The same can also be repeated for other tasks such as semantic and instance segmentation. Further, our new diagnosis tool can be employed to pinpoint weaknesses in other object detection models.

{\small
\bibliographystyle{ieee_fullname}
\bibliography{egbib}
}

\vspace{50pt}

{\bf \large
 \noindent Please download the supplementary material from here:} 

\url{https://drive.google.com/open?id=1xYUvDX_m9IEDBobRHr-KJ_pQGGJ2T-_D}

\end{document}